\documentclass[11pt,a4paper]{article}
\usepackage[nohyperref]{acl2017}
\usepackage{times}
\usepackage{latexsym}
\usepackage{amsmath}
\usepackage{amssymb}

\usepackage{multirow}
\usepackage{url}
\usepackage{todonotes}
\usepackage{gnuplot-lua-tikz}

\usepackage{xcolor}
\usepackage[normalem]{ulem}
\definecolor{darkgreen}{rgb}{0.0, 0.5, 0.0}
\definecolor{darkblue}{rgb}{0.0, 0.0, 0.8}
\definecolor{darkred}{rgb}{0.5, 0.0, 0.0}

\newcommand{\R}[2]{#1 \tiny $\pm$ \small #2}

\aclfinalcopy % Uncomment this line for the final submission
\def\aclpaperid{3580} %  Enter the acl Paper ID here

\title{Attention Strategies for Multi-Source Sequence-to-Sequence Learning}

\author{Jind\v rich Libovick\' y \and Jind\v rich Helcl \\
  Charles University, Faculty of Mathematics and Physics \\
  Institute of Formal and Applied Linguistics \\
  Malostransk\' e n\' am\v est\' i 25, 118 00 Prague, Czech Republic \\
  {\tt \{libovicky, helcl\}@ufal.mff.cuni.cz} \\}

\date{}

\begin{document}
\maketitle
\begin{abstract}

Modeling attention in neural multi-source sequence-to-sequence learning remains
a relatively unexplored area, despite its usefulness in tasks that incorporate
multiple source languages or modalities.
We propose two novel approaches to combine the outputs of attention mechanisms
over each source sequence, \emph{flat} and \emph{hierarchical}.
We compare the proposed methods with existing techniques and present results of
systematic evaluation of those methods on the WMT16 Multimodal Translation and
Automatic Post-editing tasks.
We show that the proposed methods achieve competitive results on both tasks.

\end{abstract}

% =============================================================================
\section{Introduction}
\label{sec:introduction}
% =============================================================================

Sequence-to-sequence (S2S) learning with attention mechanism recently became
the most successful paradigm with state-of-the-art results in machine
translation (MT)~\citep{bahdanau2015neural,sennrich2016wmt}, image
captioning~\citep{xu2015show,lu2016knowing}, text
summarization~\citep{rush2015summarization} and other NLP tasks.

All of the above applications of S2S learning make use of a single encoder.
Depending on the modality, it can be either a recurrent neural network (RNN)
for textual input data, or a convolutional network for images.

In this work, we focus on a special case of S2S learning with multiple input
sequences of possibly different modalities and a single output-generating
recurrent decoder. We explore various strategies the decoder can employ to
attend to the hidden states of the individual encoders.

The existing approaches to this problem do not explicitly model different
importance of the inputs to the decoder~\citep{firat2016multi,zoph2016multi}.
In multimodal MT (MMT), where an image and its caption are
on the input, we might expect the caption to be the primary source of
information, whereas the image itself would only play a role in output
disambiguation. In automatic post-editing (APE), where a sentence in a source
language and its automatically generated translation are on the input, we might
want to attend to the source text only in case the model decides that there is
an error in the translation.

We propose two interpretable attention strategies that take into account the
roles of the individual source sequences explicitly---flat and hierarchical
attention combination.

This paper is organized as follows: In Section~\ref{sec:s2c}, we review the
attention mechanism in single-source S2S learning.
Section~\ref{sec:combination} introduces new attention combination strategies.
In Section~\ref{sec:experiments}, we evaluate the proposed models on the MMT
and APE tasks.  We summarize the related work in Section~\ref{sec:related}, and
conclude in Section~\ref{sec:conclusion}.

% =============================================================================
\section{Attentive S2S Learning}
\label{sec:s2c}
% =============================================================================

The attention mechanism in S2S learning allows an RNN decoder to directly
access information about the input each time before it emits a symbol. Inspired
by content-based addressing in Neural Turing Machines~\citep{graves2014neural},
the attention mechanism estimates a probability distribution over the encoder
hidden states in each decoding step. This distribution is used for computing
the context vector---the weighted average of the encoder hidden states---as an
additional input to the decoder.

The standard attention model as described by~\citet{bahdanau2015neural} defines
the attention energies $e_{ij}$, attention distribution $\alpha_{ij}$, and the
context vector $c_i$ in $i$-th decoder step as:
\begin{gather}
    e_{ij} = v_a^\top \tanh (W_a s_i + U_a h_j), \label{eq:att_logits} \\
    \alpha_{ij} = \frac{\exp(e_{ij})}{\sum_{k=1}^{T_x}{\exp(e_{ik})}}, \label{eq:att_alphas} \\     
    c_i = \sum_{j=1}^{T_x}\alpha_{ij}h_j. \label{eq:contexts}
\end{gather}
The trainable parameters $W_a$ and $U_a$ are projection matrices that transform
the decoder and encoder states $s_i$ and $h_j$ into a common vector space and
$v_a$ is a weight vector over the dimensions of this space.  $T_x$ denotes the
length of the input sequence. For the sake of clarity, bias terms (applied
every time a vector is linearly projected using a weight matrix) are omitted.

Recently, \citet{lu2016knowing} introduced \emph{sentinel gate}, an extension
of the attentive RNN decoder with LSTM units~\citep{hochreiter1997lstm}.  We
adapt the extension for gated recurrent units (GRU)~\citep{cho2014gru}, which we
use in our experiments:
\begin{equation}
    \psi_i = \sigma(W_y y_i + W_s s_{i-1}) \label{eq:sentinel_gate}
\end{equation}
where $W_y$ and $W_s$ are trainable parameters, $y_i$ is the embedded decoder
input, and $s_{i-1}$ is the previous decoder state.

Analogically to Equation~\ref{eq:att_logits}, we compute a scalar energy term
for the sentinel:
\begin{equation} e_{\psi_i} = v_a^\top \tanh\left( W_a s_i + U_a^{(\psi)} (\psi_i
    \odot s_i) \right) \label{eq:sentinel_energy}
\end{equation}
where $W_a$, $U_a^{(\psi)}$ are the projection matrices, $v_a$ is the weight vector,
and $\psi_i \odot s_i$ is the sentinel vector. Note that the sentinel energy
term does not depend on any hidden state of any encoder.  The sentinel vector
is projected to the same vector space as the encoder state $h_j$ in
Equation~\ref{eq:att_logits}.  The term $e_{\psi_i}$ is added as an extra
attention energy term to Equation~\ref{eq:att_alphas} and the sentinel vector
$\psi_i \odot s_i$ is used as the corresponding vector in the summation in
Equation~\ref{eq:contexts}.

This technique should allow the decoder to choose whether to attend to the
encoder or to focus on its own state and act more like a language model. This
can be beneficial if the encoder does not contain much relevant information for
the current decoding step.

%Besides the sentinel vector, various other modifications to the original
%attention model have been developed. The attention mechanism can be enriched
%by explicit modeling of source coverage and
%fertility~\citep{tu2016modeling,feng2016improving}, trained to mimic GIZA++
%word alignment~\citep{alkhouli2016alignment}, or even replaced with a
%different architecture such as an additional RNN~\citep{zhang2016recurrent}.
%For simplicity, we consider only the vanilla attention model with an optional
%sentinel mechanism in the rest of the paper.

% =============================================================================
\section{Attention Combination}
\label{sec:combination}
% =============================================================================

In S2S models with multiple encoders, the decoder needs to be able to combine
the attention information collected from the encoders.

A widely adopted technique for combining multiple attention models in a decoder
is concatenation of the context vectors $c_i^{(1)}, \ldots,
c_i^{(N)}$~\citep{zoph2016multi,firat2016multi}. As mentioned in
Section~\ref{sec:introduction}, this setting forces the model to attend to each
encoder independently and lets the attention combination to be resolved
implicitly in the subsequent network layers.

In this section, we propose two alternative strategies of combining attentions
from multiple encoders. We either let the decoder learn the $\mathbf{\alpha}_i$
distribution jointly over all encoder hidden states (\emph{flat} attention combination) or factorize the distribution over individual encoders
(\emph{hierarchical} combination).

Both of the alternatives allow us to explicitly compute distribution over the
encoders and thus interpret how much attention is paid to each encoder at every
decoding step.

% -----------------------------------------------------------------------------
\subsection{Flat Attention Combination}
% -----------------------------------------------------------------------------

Flat attention combination projects the hidden states of all encoders into a
shared space and then computes an arbitrary distribution over the projections.  The
difference between the concatenation of the context vectors and the flat attention
combination is that the $\mathbf{\alpha}_{i}$ coefficients are computed jointly
for all encoders:
\begin{equation}
    \alpha_{ij}^{(k)} = \frac{\exp(e_{ij}^{(k)})} {\sum_{n=1}^N
    \sum_{m=1}^{T_x^{(n)}}{\exp \left (e_{im}^{(n)} \right)}}
\end{equation}
where $T_x^{(n)}$ is the length of the input sequence of the $n$-th encoder and
${e}_{ij}^{(k)}$ is the attention energy of the $j$-th state of the $k$-th
encoder in the $i$-th decoding step.
%
%\begin{gather}
%
% \hat{e}_i = e_i^{(1)} \oplus e_i^{(2)} \oplus \ldots \oplus e_i^{(N)} \\
%
%e_{ij}^{(k)} = v_a^\top \tanh \left( W_a s_i + U_a^{(k)} h_j^{(k)} \right)
%
%\end{gather}
%
These attention energies are computed as in Equation~\ref{eq:att_logits}. The
parameters $v_a$ and $W_a$ are shared among the encoders, and $U_a$ is
different for each encoder and serves as an encoder-specific projection of
hidden states into a common vector space.

The states of the individual encoders occupy different vector spaces and
can have a different dimensionality, therefore the context vector cannot be computed
as their weighted sum. We project them into a single space using
linear projections:
\begin{gather}
%
%\hat{h}_j = U_c^{(1)} h_j^{(1)} \oplus \ldots \oplus U_c^{(N)} h_j^{(N)}
    %\label{eq:proj-states} \\
%
%c_i = \sum_{j=1}^{\hat{T_x}}\alpha_{ij}\hat{h}_j \label{eq:flat-context}
%
    c_i = \sum_{k=1}^N \sum_{j=1}^{T_x^{(k)}}\alpha_{ij}^{(k)}U_c^{(k)} h_j^{(k)} 
    \label{eq:flat-context}
\end{gather}
where $U_c^{(k)}$ are additional trainable parameters.

The matrices $U_c^{(k)}$ project the hidden states into a common
vector space. This raises a question whether this space
can be the same as the one that is projected into in the energy 
computation using matrices $U_a^{(k)}$ in 
Equation~\ref{eq:att_logits}, i.e., whether 
$U_c^{(k)} = U_a^{(k)}$. In our experiments, we explore both options.
We also try both adding and not adding the sentinel
$\alpha_i^{(\psi)}U_c^{(\psi)}(\psi_i \odot s_i)$ to the  context vector.

% -----------------------------------------------------------------------------
\subsection{Hierarchical Attention Combination}
% -----------------------------------------------------------------------------

The hierarchical attention combination model computes every context vector
independently,
similarly to the concatenation approach. Instead of concatenation, a second
attention mechanism is constructed over the context vectors.

We divide the computation of the attention distribution into two steps:
First, we compute the context vector for each encoder independently using
Equation~\ref{eq:contexts}. Second, we project the context vectors 
(and optionally the sentinel) into a common space (Equation~\ref{eq:hier_ener}), we compute
another distribution over the projected context vectors (Equation~\ref{eq:hier_distr}) and their corresponding
weighted average (Equation~\ref{eq:hier_context}):
\begin{gather}
    e_{i}^{(k)} = v_b^\top \tanh (W_b s_{i} + U_b^{(k)} c_i^{(k)}), \label{eq:hier_ener}\\
	\beta_i^{(k)} = \frac{\exp(e_i^{(k)})} { \sum_{n=1}^{N} \exp(e_i^{(n)}) }, \label{eq:hier_distr} \\ %
	c_i = \sum_{k=1}^N \beta_i^{(k)} U_c^{(k)} c_i^{(k)} \label{eq:hier_context}
   %= \sum_{k=1}^N \beta_i^{(k)} U_c^{(k)} \sum_{j=1} \alpha_{ij}^{(k)} h_j^{(k)}
\end{gather}
where $c_i^{(k)}$ is the context vector of the $k$-th encoder, additional
trainable parameters $v_b$ and $W_b$ are shared for all encoders, and $U_b^{(k)}$ and $U_c^{(k)}$
are encoder-specific projection matrices, that can be set equal and shared, 
similarly to the case of flat attention combination.

% =============================================================================
\section{Experiments}
\label{sec:experiments}
% =============================================================================

We evaluate the attention combination strategies presented in Section~\ref{sec:combination}
on the tasks of multimodal translation (Section~\ref{sec:mmmt}) and automatic post-editing (Section~\ref{sec:ape}).

The models were implemented using the Neural Monkey
sequence-to-sequence learning
toolkit~\citep{helcl2017neural}.\footnote{\url{http://github.com/ufal/neuralmonkey}}
%\footnote{The trained models can be downloaded from \\ \url{http://ufallab.ms.mff.cuni.cz/~libovicky/acl2017_att_models/}}
In both setups, we process the textual input with bidirectional GRU
network~\citep{cho2014gru} with 300 units in the hidden state in each direction 
and 300 units in embeddings. For the attention projection space, we use 500 hidden units.
We optimize the network to minimize the output 
cross-entropy using the Adam algorithm~\citep{kingma2015adam} with learning rate 
$10^{-4}$.

% -----------------------------------------------------------------------------
\subsection{Multimodal Translation}
\label{sec:mmmt}
% -----------------------------------------------------------------------------

% % % % % % % % % % % % % % % % % % % % % % % % % % % % % % % % % % % % % % % %
\begin{figure}

	\centering
    \input{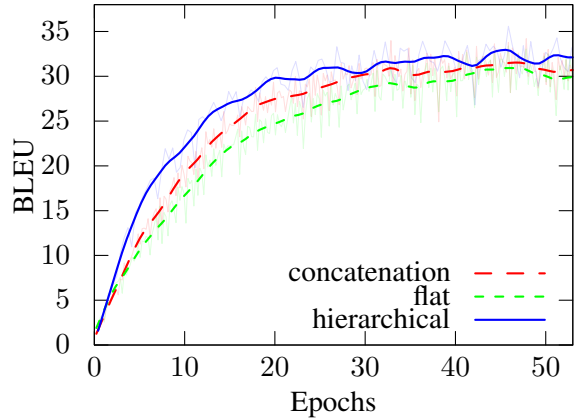}

    \caption{Learning curves on validation data for context vector
        concatenation (blue), flat (green) and hierarchical (red) attention
        combination without sentinel and without sharing the projection
        matrices.}
        
     \label{fig:learningcurves}

\end{figure}
% % % % % % % % % % % % % % % % % % % % % % % % % % % % % % % % % % % % % % % %

The goal of multimodal translation~\citep{specia2016shared} is to generate 
target-language image captions given both the image and its caption in the
source language. 

We train and evaluate the model on the Multi30k 
dataset~\citep{elliott2016multi30k}. It consists of 29,000 training
instances (images together with English captions and their German 
translations), 
1,014 validation instances, and 1,000 test instances. The results are
evaluated using the BLEU~\citep{papineni2002bleu} and 
METEOR~\citep{denkowski2011meteor}.

In our model, the visual input is processed with a pre-trained 
VGG~16 network \citep{simonyan2014vgg} without further fine-tuning. Attention 
distribution over the visual input is computed from the last convolutional layer 
of the network. The decoder is an RNN with 500 conditional GRU 
units~\citep{firat2016cgru} in the recurrent layer.
We use byte-pair encoding~\citep{sennrich2016neural} with a vocabulary 
of 20,000 subword units shared between the textual encoder and the decoder.

The results of our experiments in multimodal MT are shown in Table~\ref{tab:mmmt}. We achieved the best results using the hierarchical attention combination
without the sentinel mechanism, which also showed the fastest convergence.
The flat combination strategy achieves similar results eventually. Sharing the projections
for energy and context vector computation does not improve over the concatenation baseline
and slows the training almost prohibitively.
Multimodal models were not able to surpass the textual baseline (BLEU 33.0).

Using the conditional GRU units brought an improvement of about 1.5 BLEU points
on average, with the exception of the concatenation scenario where
the performance dropped by almost 5 BLEU points. 
We hypothesize this is caused by the fact the model has to learn the
implicit attention combination on multiple places -- once in the 
output projection and three times inside the conditional GRU
unit~\citep[Equations 10-12]{firat2016cgru}.
We thus report the scores of the introduced attention combination techniques
trained with conditional GRU units and compare them with the concatenation
baseline trained with plain GRU units.

% -----------------------------------------------------------------------------
\subsection{Automatic MT Post-editing}
\label{sec:ape}
% -----------------------------------------------------------------------------

Automatic post-editing is a task of improving an automatically generated 
translation given the source sentence where the translation system is treated
as a black box.

We used the data from the WMT16 APE Task~\cite{bojar2016findings,turich2016ape},
which consists of 12,000 training, 2,000 validation, and 1,000 test sentence 
triplets from the IT domain. Each triplet contains an English source sentence,
an automatically generated German translation of the source sentence,
and a manually post-edited German sentence as a reference. In case of this 
dataset, the MT outputs 
are almost perfect in and only little effort was
required to post-edit the sentences.
The results are evaluated using the human-targeted error rate (HTER)~\citep{snover2006study} and BLEU score~\citep{papineni2002bleu}.

Following \citet{libovicky2016cuni}, we encode the target sentence as 
a sequence of edit operations transforming the MT output into the reference.
By this technique, we prevent the model from paraphrasing the input sentences.
The decoder is a GRU network with 300 hidden units. Unlike in the MMT setup (Section~\ref{sec:mmmt}),
we do not use the conditional GRU
because it is prone to overfitting on the small dataset we work with.

The models were able to slightly, but significantly improve over the baseline -- leaving 
the MT output as is (HTER 24.8). The differences between the attention combination
strategies are not significant.

% % % % % % % % % % % % % % % % % % % % % % % % % % % % % % % % % % % % % % %
\begin{table}

\centering
\scalebox{.85}{
\begin{tabular}{l|c|c||c|c||c|c}
\multirow{2}{*}{} &
\multirow{2}{*}{\rotatebox{90}{share}} &
\multirow{2}{*}{\rotatebox{90}{sent.}} &
\multicolumn{2}{c||}{MMT} & \multicolumn{2}{c}{APE} \\
& & & B\scalebox{.8}{LEU} & M\scalebox{.8}{ETEOR} & 
      B\scalebox{.8}{LEU} & H\scalebox{.8}{TER} \\ \hline \hline
%\multicolumn{3}{l||}{baseline} 
%                              & 32.4 & 49.3  & ??.? & ??.? \\ \hline
\multicolumn{3}{l||}{concat.}
                              & \R{31.4}{.8} & \R{48.0}{.7}  & \R{62.3}{.5} & \R{24.4}{.4} \\ \hline
\multirow{4}{*}{\rotatebox{90}{flat}}
    & $\times$ & $\times$     & \R{30.2}{.8} & \R{46.5}{.7} & \R{62.6}{.5} & \R{24.2}{.4} \\
    & $\times$ & \checkmark   & \R{29.3}{.8} & \R{45.4}{.7} & \R{62.3}{.5} & \R{24.3}{.4} \\
    & \checkmark & $\times$   & \R{30.9}{.8} & \R{47.1}{.7} & \R{62.4}{.6} & \R{24.4}{.4} \\
    & \checkmark & \checkmark & \R{29.4}{.8} & \R{46.9}{.7} & \R{62.5}{.6} & \R{24.2}{.4} \\ \hline
\multirow{4}{*}{\rotatebox{90}{hierarchical~}}
    & $\times$ & $\times$     & \textbf{\R{32.1}{.8}} & \textbf{\R{49.1}{.7}} & \R{62.3}{.5} & \R{24.1}{.4} \\
    & $\times$ & \checkmark   & \R{28.1}{.8} & \R{45.5}{.7} & \R{62.6}{.6} & \R{24.1}{.4}\\
    & \checkmark & $\times$   & \R{26.1}{.7} & \R{42.4}{.7} & \R{62.4}{.5} & \R{24.3}{.4}  \\
    & \checkmark & \checkmark & \R{22.0}{.7} & \R{38.5}{.6} & \R{62.5}{.5} & \R{24.1}{.4} \\
\end{tabular}}

\caption{Results of our experiments on the test sets of Multi30k dataset and
the APE dataset. The column `share' denotes whether the projection matrix
is shared for energies and context vector computation, `sent.' indicates whether the sentinel
vector has been used or not.}

\label{tab:mmmt}

\end{table}
% % % % % % % % % % % % % % % % % % % % % % % % % % % % % % % % % % % % % % %

% % % % % % % % % % % % % % % % % % % % % % % % % % % % % % % % % % % % % % %
\begin{figure}[t!]
    \begin{center}
    \includegraphics[width=.3\textwidth]{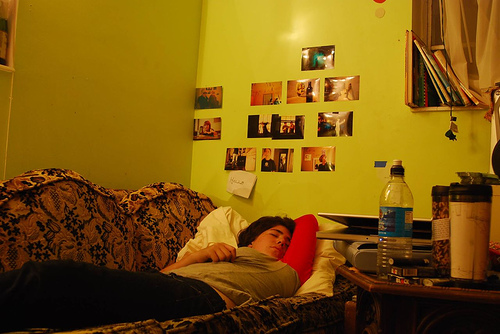}
    \end{center}
	\noindent
    \textbf{Source:}  a man sleeping in a green room on a couch .
    
    \noindent
    \textbf{Reference:} ein Mann schl\" aft in einem gr\" unen Raum auf einem Sofa .
    
    \vspace{2pt}
    \noindent
    \textbf{Output with attention:}
    
    \includegraphics[width=.4\textwidth]{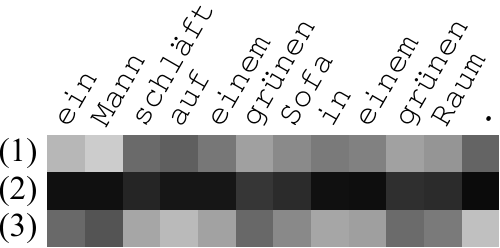} \\
    \centering (1) source, (2) image, (3) sentinel

    \caption{Visualization of hierarchical attention in MMT. Each column in the diagram
    corresponds to the weights of the encoders and sentinel.     
    Note that the despite the overall low importance of the image encoder, it gets
    activated for the content words.}

\end{figure}
% % % % % % % % % % % % % % % % % % % % % % % % % % % % % % % % % % % % % % %

% =============================================================================
\section{Related Work}
\label{sec:related}
% =============================================================================

Attempts to use S2S models for APE are relatively
rare~\citep{bojar2016findings}. \citet{niehues2016pretranslation} concatenate
both inputs into one long sequence, which forces the encoder to be able to work
with both source and target language.  Their attention is then similar to our
flat combination strategy; however, it can only be used for sequential data.

The best system from the WMT'16 competition~\citep{marcin2016ape} trains two
separate S2S models, one translating from MT output to post-edited targets and the second one
from source sentences to post-edited targets. The decoders average their
output distributions similarly to decoder ensembling.
The biggest source of improvement in
this state-of-the-art posteditor came from additional training data generation,
rather than from changes in the network architecture.

\citet{caglayan2016multimodality} used an architecture very similar to ours
for multimodal translation. They made a strong assumption that the network
can be trained in such a way that the hidden states of the encoder
and the convolutional network occupy the same vector space and thus sum
the context vectors from both modalities. In this way, their multimodal MT system 
(BLEU 27.82) remained far bellow  the text-only setup (BLEU 32.50).

New state-of-the-art results on the Multi30k dataset were achieved very recently by
\citet{calixtoL2017incorporating}. The best-performing architecture uses the
last fully-connected layer of VGG-19 network~\citep{simonyan2014vgg} as decoder
initialization and only attends to the text encoder hidden states.
With a stronger monomodal baseline (BLEU 33.7), their multimodal model achieved
a BLEU score of 37.1. Similarly to \citet{niehues2016pretranslation} in the APE
task, even further improvement was achieved by synthetically extending the
dataset.

% =============================================================================
\section{Conclusions}
\label{sec:conclusion}
% =============================================================================

We introduced two new strategies of combining attention in a multi-source
sequence-to-sequence setup. Both methods are based on computing a
joint distribution over hidden states of all encoders.

We conducted experiments with the proposed strategies on multimodal translation 
and automatic post-editing tasks, and we showed that the flat and hierarchical
attention combination can be applied to these tasks with maintaining competitive
score to previously used techniques.

Unlike the simple context vector concatenation, the introduced combination strategies
can be used with the conditional GRU units in the decoder. On top of that, the hierarchical 
combination strategy exhibits faster learning than than the other strategies.

\section*{Acknowledgments}

We would like to thank Ond\v rej Du\v sek, Rudolf Rosa, Pavel Pecina, and Ond\v rej Bojar 
for a fruitful discussions and comments on the draft of the paper.

This research has been funded by the Czech Science Foundation grant 
no. P103/12/G084, the EU grant no. H2020-ICT-2014-1-645452 (QT21),
and Charles University grant no. 52315/2014 and SVV project no. 260 453.
This work has been using language resources developed and/or stored and/or distributed
by the LINDAT-Clarin project of the Ministry of Education of the Czech Republic (project
LM2010013).

% The acknowledgments should go immediately before the references.  Do
% not number the acknowledgments section. Do not include this section
% when submitting your paper for review.

% % include your own bib file like this:
% %\bibliographystyle{acl}
% %\bibliography{acl2017}
\bibliography{acl2017}

\begin{thebibliography}{}
\expandafter\ifx\csname natexlab\endcsname\relax\def\natexlab#1{#1}\fi

\bibitem[{Bahdanau et~al.(2014)Bahdanau, Cho, and Bengio}]{bahdanau2015neural}
Dzmitry Bahdanau, Kyunghyun Cho, and Yoshua Bengio. 2014.
\newblock \href{http://arxiv.org/abs/1409.0473}{Neural machine translation by
  jointly learning to align and translate}.
\newblock {\em CoRR\/} abs/1409.0473.
\newblock
  \href{http://arxiv.org/abs/1409.0473}{http://arxiv.org/abs/1409.0473}.

\bibitem[{Bojar et~al.(2016)Bojar, Chatterjee, Federmann, Graham, Haddow, Huck,
  Yepes, Koehn, Logacheva, Monz, Negri, N{\'{e}}v{\'{e}}ol, Neves, Popel, Post,
  Rubino, Scarton, Specia, Turchi, Verspoor, and Zampieri}]{bojar2016findings}
Ond{\v{r}}ej Bojar, Rajen Chatterjee, Christian Federmann, Yvette Graham, Barry
  Haddow, Matthias Huck, Antonio Yepes, Philipp Koehn, Varvara Logacheva,
  Christof Monz, Matteo Negri, Aurelie N{\'{e}}v{\'{e}}ol, Mariana Neves,
  Martin Popel, Matt Post, Raphael Rubino, Carolina Scarton, Lucia Specia,
  Marco Turchi, Karin Verspoor, and Marcos Zampieri. 2016.
\newblock Findings of the 2016 conference on machine translation ({WMT}16).
\newblock In {\em Proceedings of the First Conference on Machine Translation
  ({WMT}). Volume 2: Shared Task Papers\/}. Association for Computational
  Linguistics, Association for Computational Linguistics, Stroudsburg, {PA},
  {USA}, volume~2, pages 131--198.

\bibitem[{Caglayan et~al.(2016)Caglayan, Aransa, Wang, Masana,
  Garc\'{i}a-Mart\'{i}nez, Bougares, Barrault, and van~de
  Weijer}]{caglayan2016multimodality}
Ozan Caglayan, Walid Aransa, Yaxing Wang, Marc Masana, Mercedes
  Garc\'{i}a-Mart\'{i}nez, Fethi Bougares, Lo\"{i}c Barrault, and Joost van~de
  Weijer. 2016.
\newblock \href{http://www.aclweb.org/anthology/W16-2358}{Does multimodality
  help human and machine for translation and image captioning?}
\newblock In {\em Proceedings of the First Conference on Machine
  Translation\/}. Association for Computational Linguistics, Berlin, Germany,
  pages 627--633.
\newblock
  \href{http://www.aclweb.org/anthology/W16-2358}{http://www.aclweb.org/anthology/W16-2358}.

\bibitem[{Calixto et~al.(2017)Calixto, Liu, and
  Campbell}]{calixtoL2017incorporating}
Iacer Calixto, Qun Liu, and Nick Campbell. 2017.
\newblock \href{http://arxiv.org/abs/1701.06521}{Incorporating global visual
  features into attention-based neural machine translation}.
\newblock {\em CoRR\/} abs/1701.06521.
\newblock
  \href{http://arxiv.org/abs/1701.06521}{http://arxiv.org/abs/1701.06521}.

\bibitem[{Cho et~al.(2014)Cho, van Merrienboer, Bahdanau, and
  Bengio}]{cho2014gru}
Kyunghyun Cho, Bart van Merrienboer, Dzmitry Bahdanau, and Yoshua Bengio. 2014.
\newblock \href{http://www.aclweb.org/anthology/W14-4012}{On the properties of
  neural machine translation: Encoder--decoder approaches}.
\newblock In {\em Proceedings of SSST-8, Eighth Workshop on Syntax, Semantics
  and Structure in Statistical Translation\/}. Association for Computational
  Linguistics, Doha, Qatar, pages 103--111.
\newblock
  \href{http://www.aclweb.org/anthology/W14-4012}{http://www.aclweb.org/anthology/W14-4012}.

\bibitem[{Denkowski and Lavie(2011)}]{denkowski2011meteor}
Michael Denkowski and Alon Lavie. 2011.
\newblock \href{http://www.aclweb.org/anthology/W11-2107}{Meteor 1.3:
  {A}utomatic metric for reliable optimization and evaluation of machine
  translation systems}.
\newblock In {\em Proceedings of the Sixth Workshop on Statistical Machine
  Translation\/}. Association for Computational Linguistics, Edinburgh, United
  Kingdom, pages 85--91.
\newblock
  \href{http://www.aclweb.org/anthology/W11-2107}{http://www.aclweb.org/anthology/W11-2107}.

\bibitem[{Elliott et~al.(2016)Elliott, Frank, Sima'an, and
  Specia}]{elliott2016multi30k}
Desmond Elliott, Stella Frank, Khalil Sima'an, and Lucia Specia. 2016.
\newblock \href{http://arxiv.org/abs/1605.00459}{Multi30k: Multilingual
  english-german image descriptions}.
\newblock {\em CoRR\/} abs/1605.00459.
\newblock
  \href{http://arxiv.org/abs/1605.00459}{http://arxiv.org/abs/1605.00459}.

\bibitem[{Firat and Cho(2016)}]{firat2016cgru}
Orhan Firat and Kyunghyun Cho. 2016.
\newblock Conditional gated recurrent unit with attention mechanism.
\newblock https://github.com/nyu-dl/dl4mt-tutorial/blob/master/docs/cgru.pdf.
\newblock Published online, version {\tt adbaeea}.

\bibitem[{Firat et~al.(2016)Firat, Cho, and Bengio}]{firat2016multi}
Orhan Firat, Kyunghyun Cho, and Yoshua Bengio. 2016.
\newblock \href{http://www.aclweb.org/anthology/N16-1101}{Multi-way,
  multilingual neural machine translation with a shared attention mechanism}.
\newblock In {\em Proceedings of the 2016 Conference of the North American
  Chapter of the Association for Computational Linguistics: Human Language
  Technologies\/}. Association for Computational Linguistics, San Diego, CA,
  USA, pages 866--875.
\newblock
  \href{http://www.aclweb.org/anthology/N16-1101}{http://www.aclweb.org/anthology/N16-1101}.

\bibitem[{Graves et~al.(2014)Graves, Wayne, and Danihelka}]{graves2014neural}
Alex Graves, Greg Wayne, and Ivo Danihelka. 2014.
\newblock \href{http://arxiv.org/abs/1410.5401}{Neural turing machines}.
\newblock {\em CoRR\/} abs/1410.5401.
\newblock
  \href{http://arxiv.org/abs/1410.5401}{http://arxiv.org/abs/1410.5401}.

\bibitem[{Helcl and Libovick{\'{y}}(2017)}]{helcl2017neural}
Jind{\v{r}}ich Helcl and Jind{\v{r}}ich Libovick{\'{y}}. 2017.
\newblock \href{https://doi.org/10.1515/pralin-2017-0001}{Neural monkey: An
  open-source tool for sequence learning}.
\newblock {\em The Prague Bulletin of Mathematical Linguistics\/} (107):5--17.
\newblock
  \href{https://doi.org/10.1515/pralin-2017-0001}{https://doi.org/10.1515/pralin-2017-0001}.

\bibitem[{Hochreiter and Schmidhuber(1997)}]{hochreiter1997lstm}
Sepp Hochreiter and J\"{u}rgen Schmidhuber. 1997.
\newblock \href{https://doi.org/10.1162/neco.1997.9.8.1735}{Long short-term
  memory}.
\newblock {\em Neural Comput.\/} 9:1735--1780.
\newblock
  \href{https://doi.org/10.1162/neco.1997.9.8.1735}{https://doi.org/10.1162/neco.1997.9.8.1735}.

\bibitem[{Junczys-Dowmunt and Grundkiewicz(2016)}]{marcin2016ape}
Marcin Junczys-Dowmunt and Roman Grundkiewicz. 2016.
\newblock \href{http://www.aclweb.org/anthology/W/W16/W16-2378}{Log-linear
  combinations of monolingual and bilingual neural machine translation models
  for automatic post-editing}.
\newblock In {\em Proceedings of the First Conference on Machine
  Translation\/}. Association for Computational Linguistics, Berlin, Germany,
  pages 751--758.
\newblock
  \href{http://www.aclweb.org/anthology/W/W16/W16-2378}{http://www.aclweb.org/anthology/W/W16/W16-2378}.

\bibitem[{Kingma and Ba(2014)}]{kingma2015adam}
Diederik~P. Kingma and Jimmy Ba. 2014.
\newblock \href{http://arxiv.org/abs/1412.6980}{Adam: {A} method for stochastic
  optimization}.
\newblock {\em CoRR\/} abs/1412.6980.
\newblock
  \href{http://arxiv.org/abs/1412.6980}{http://arxiv.org/abs/1412.6980}.

\bibitem[{Libovick\'{y} et~al.(2016)Libovick\'{y}, Helcl, Tlust\'{y}, Bojar,
  and Pecina}]{libovicky2016cuni}
Jind\v{r}ich Libovick\'{y}, Jind\v{r}ich Helcl, Marek Tlust\'{y}, Ond\v{r}ej
  Bojar, and Pavel Pecina. 2016.
\newblock \href{http://www.aclweb.org/anthology/W/W16/W16-2361}{{CUNI} system
  for {WMT16} automatic post-editing and multimodal translation tasks}.
\newblock In {\em Proceedings of the First Conference on Machine
  Translation\/}. Association for Computational Linguistics, Berlin, Germany,
  pages 646--654.
\newblock
  \href{http://www.aclweb.org/anthology/W/W16/W16-2361}{http://www.aclweb.org/anthology/W/W16/W16-2361}.

\bibitem[{Lu et~al.(2016)Lu, Xiong, Parikh, and Socher}]{lu2016knowing}
Jiasen Lu, Caiming Xiong, Devi Parikh, and Richard Socher. 2016.
\newblock \href{http://arxiv.org/abs/1612.01887}{Knowing when to look:
  {A}daptive attention via a visual sentinel for image captioning}.
\newblock {\em CoRR\/} abs/1612.01887.
\newblock
  \href{http://arxiv.org/abs/1612.01887}{http://arxiv.org/abs/1612.01887}.

\bibitem[{Niehues et~al.(2016)Niehues, Cho, Ha, and
  Waibel}]{niehues2016pretranslation}
Jan Niehues, Eunah Cho, Thanh{-}Le Ha, and Alex Waibel. 2016.
\newblock \href{http://arxiv.org/abs/1610.05243}{Pre-translation for neural
  machine translation}.
\newblock {\em CoRR\/} abs/1610.05243.
\newblock
  \href{http://arxiv.org/abs/1610.05243}{http://arxiv.org/abs/1610.05243}.

\bibitem[{Papineni et~al.(2002)Papineni, Roukos, Ward, and
  Zhu}]{papineni2002bleu}
Kishore Papineni, Salim Roukos, Todd Ward, and Wei-Jing Zhu. 2002.
\newblock \href{https://doi.org/10.3115/1073083.1073135}{Bleu: a method for
  automatic evaluation of machine translation}.
\newblock In {\em Proceedings of 40th Annual Meeting of the Association for
  Computational Linguistics\/}. Association for Computational Linguistics,
  Philadelphia, Pennsylvania, USA, pages 311--318.
\newblock
  \href{https://doi.org/10.3115/1073083.1073135}{https://doi.org/10.3115/1073083.1073135}.

\bibitem[{Rush et~al.(2015)Rush, Chopra, and Weston}]{rush2015summarization}
Alexander~M. Rush, Sumit Chopra, and Jason Weston. 2015.
\newblock \href{https://aclweb.org/anthology/D/D15/D15-1044}{A neural attention
  model for abstractive sentence summarization}.
\newblock In {\em Proceedings of the 2015 Conference on Empirical Methods in
  Natural Language Processing\/}. Association for Computational Linguistics,
  Lisbon, Portugal, pages 379--389.
\newblock
  \href{https://aclweb.org/anthology/D/D15/D15-1044}{https://aclweb.org/anthology/D/D15/D15-1044}.

\bibitem[{Sennrich et~al.(2016{\natexlab{a}})Sennrich, Haddow, and
  Birch}]{sennrich2016wmt}
Rico Sennrich, Barry Haddow, and Alexandra Birch. 2016{\natexlab{a}}.
\newblock \href{http://www.aclweb.org/anthology/W/W16/W16-2323}{Edinburgh
  neural machine translation systems for {WMT} 16}.
\newblock In {\em Proceedings of the First Conference on Machine
  Translation\/}. Association for Computational Linguistics, Berlin, Germany,
  pages 371--376.
\newblock
  \href{http://www.aclweb.org/anthology/W/W16/W16-2323}{http://www.aclweb.org/anthology/W/W16/W16-2323}.

\bibitem[{Sennrich et~al.(2016{\natexlab{b}})Sennrich, Haddow, and
  Birch}]{sennrich2016neural}
Rico Sennrich, Barry Haddow, and Alexandra Birch. 2016{\natexlab{b}}.
\newblock \href{http://www.aclweb.org/anthology/P16-1162}{Neural machine
  translation of rare words with subword units}.
\newblock In {\em Proceedings of the 54th Annual Meeting of the Association for
  Computational Linguistics (Volume 1: Long Papers)\/}. Association for
  Computational Linguistics, Berlin, Germany, pages 1715--1725.
\newblock
  \href{http://www.aclweb.org/anthology/P16-1162}{http://www.aclweb.org/anthology/P16-1162}.

\bibitem[{Simonyan and Zisserman(2014)}]{simonyan2014vgg}
Karen Simonyan and Andrew Zisserman. 2014.
\newblock \href{http://arxiv.org/abs/1409.1556}{Very deep convolutional
  networks for large-scale image recognition}.
\newblock {\em CoRR\/} abs/1409.1556.
\newblock
  \href{http://arxiv.org/abs/1409.1556}{http://arxiv.org/abs/1409.1556}.

\bibitem[{Snover et~al.(2006)Snover, Dorr, Schwartz, Micciulla, and
  Makhoul}]{snover2006study}
Matthew Snover, Bonnie Dorr, Richard Schwartz, Linnea Micciulla, and John
  Makhoul. 2006.
\newblock A study of translation edit rate with targeted human annotation.
\newblock In {\em Proceedings of association for machine translation in the
  Americas\/}. volume 200.

\bibitem[{Specia et~al.(2016)Specia, Frank, Sima'an, and
  Elliott}]{specia2016shared}
Lucia Specia, Stella Frank, Khalil Sima'an, and Desmond Elliott. 2016.
\newblock \href{http://www.aclweb.org/anthology/W16-2346}{A shared task on
  multimodal machine translation and crosslingual image description}.
\newblock In {\em Proceedings of the First Conference on Machine
  Translation\/}. Association for Computational Linguistics, Berlin, Germany,
  pages 543--553.
\newblock
  \href{http://www.aclweb.org/anthology/W16-2346}{http://www.aclweb.org/anthology/W16-2346}.

\bibitem[{Turchi et~al.(2016)Turchi, Chatterjee, and Negri}]{turich2016ape}
Marco Turchi, Rajen Chatterjee, and Matteo Negri. 2016.
\newblock \href{http://hdl.handle.net/11372/LRT-1632}{{WMT16} {APE} shared task
  data}.
\newblock {LINDAT}/{CLARIN} digital library at the Institute of Formal and
  Applied Linguistics, Charles University in Prague.
\newblock
  \href{http://hdl.handle.net/11372/LRT-1632}{http://hdl.handle.net/11372/LRT-1632}.

\bibitem[{Xu et~al.(2015)Xu, Ba, Kiros, Cho, Courville, Salakhudinov, Zemel,
  and Bengio}]{xu2015show}
Kelvin Xu, Jimmy Ba, Ryan Kiros, Kyunghyun Cho, Aaron Courville, Ruslan
  Salakhudinov, Rich Zemel, and Yoshua Bengio. 2015.
\newblock \href{http://jmlr.org/proceedings/papers/v37/xuc15.pdf}{Show, attend
  and tell: {N}eural image caption generation with visual attention}.
\newblock In David Blei and Francis Bach, editors, {\em Proceedings of the 32nd
  International Conference on Machine Learning (ICML-15)\/}. JMLR Workshop and
  Conference Proceedings, Lille, France, pages 2048--2057.
\newblock
  \href{http://jmlr.org/proceedings/papers/v37/xuc15.pdf}{http://jmlr.org/proceedings/papers/v37/xuc15.pdf}.

\bibitem[{Zoph and Knight(2016)}]{zoph2016multi}
Barret Zoph and Kevin Knight. 2016.
\newblock \href{http://www.aclweb.org/anthology/N16-1004}{Multi-source neural
  translation}.
\newblock In {\em Proceedings of the 2016 Conference of the North American
  Chapter of the Association for Computational Linguistics: Human Language
  Technologies\/}. Association for Computational Linguistics, San Diego, CA,
  USA, pages 30--34.
\newblock
  \href{http://www.aclweb.org/anthology/N16-1004}{http://www.aclweb.org/anthology/N16-1004}.

\end{thebibliography}
\bibliographystyle{acl_natbib}

% \appendix

% \section{Supplemental Material}
% \label{sec:supplemental}

\textit{}\end{document}